\documentclass[runningheads]{llncs}

\usepackage{eccv}

\usepackage{eccvabbrv}

\usepackage{graphicx}
\usepackage{booktabs}

\usepackage[accsupp]{axessibility}  

\usepackage[breaklinks,colorlinks,citecolor=eccvblue]{hyperref}

\usepackage{orcidlink}
\usepackage{longtable}
\usepackage{lipsum}
\newcommand\blfootnote[1]{%
  \begingroup
  \renewcommand\thefootnote{}\footnote{#1}%
  \addtocounter{footnote}{-1}%
  \endgroup
}
\begin{document}

\title{GRITv2: Efficient and Light-weight Social Relation Recognition} 

\titlerunning{GRITv2}

\author{N K Sagar Reddy\and
Neeraj Kasera \and
Avinash Thakur}

\authorrunning{Sagar Reddy et al.}

\institute{OPPO Mobiles R\&D, India
}

\maketitle

\begin{abstract}


Our research focuses on the analysis and improvement of the Graph-based Relation Inference Transformer (GRIT), which serves as an important benchmark in the field. We conduct a comprehensive ablation study using the PISC-fine dataset, to find and explore improvement in efficiency and performance of GRITv2. Our research has provided a new state-of-the-art relation recognition model on the PISC relation dataset. We introduce several features in the GRIT model and analyse our new benchmarks in two versions: GRITv2-L (large) and GRITv2-S (small). Our proposed GRITv2-L surpasses existing methods on relation recognition and the GRITv2-S is within 2\% performance gap of GRITv2-L, which has only 0.0625x the model size and parameters of GRITv2-L. Furthermore, we also address the need for model compression, an area crucial for deploying efficient models on resource-constrained platforms. By applying quantization techniques, we efficiently reduced the GRITv2-S size to 22MB and deployed it on the flagship OnePlus 12 mobile which still surpasses the PISC-fine benchmarks in performance, highlighting the practical viability and improved efficiency of our model on mobile devices.\blfootnote{Arxiv Preprint}


\keywords{Relation Recognition, Quantization, Mobile Deployment.}
\end{abstract}

\section{Introduction}
\label{sec:intro}
In the realm of artificial intelligence, the ability to comprehend and interpret human relations stands as a pivotal task. Human relations encapsulate various details like intricate human interactions, emotions, and social dynamics. Accurate classification of human relations within images not only facilitates a deeper understanding of human behavior but also empowers AI systems to build refined knowledge about each individual.

Understanding and recognizing human relations is also crucial in mobile devices for several reasons. Mobile devices have become integral parts of our daily lives, deeply intertwined with how we communicate, work, entertain ourselves, and manage our personal affairs. Recognizing human relations allows mobile devices to personalize content, services, and recommendations based on individual preferences, behaviors, and social connections. Social networks also play a central role in facilitating social connections and communication. Human relation recognition helps these platforms create features that foster meaningful connections, support collaboration, and strengthen social bonds among users.

People in Social Context (PISC) dataset is a majorly used dataset in the human relation recognition task, introduced by \cite{PISC_dualglance}. This dataset consists of 6 fine-level relation labels (PISC-F) and 3 coarse-level relation labels (PISC-C). In the recent years, multiple state-of-the-art models \cite{li_et_al,isl,mt_srr} have been proposed to improve the performance benchmarks for the PISC dataset.

Our work is based on GRIT\cite{grit} proposed in 2022. We improve the model and propose GRITv2, by analysing the PISC dataset and performing model compression studies on the model. We succeed in creating a state-of-the-art model surpassing existing benchmarks and also efficiently deploy it on a mobile device with minimal loss of performance.

\begin{table}[ht]
\caption{Visual results of GRITv2-L on PISC-Fine dataset\cite{PISC_dualglance}.}
\centering
\begin{tabular}{cc}
\textbf{PISC-Fine demo split} & \textbf{Relation classification} \\
\includegraphics[scale=0.075]{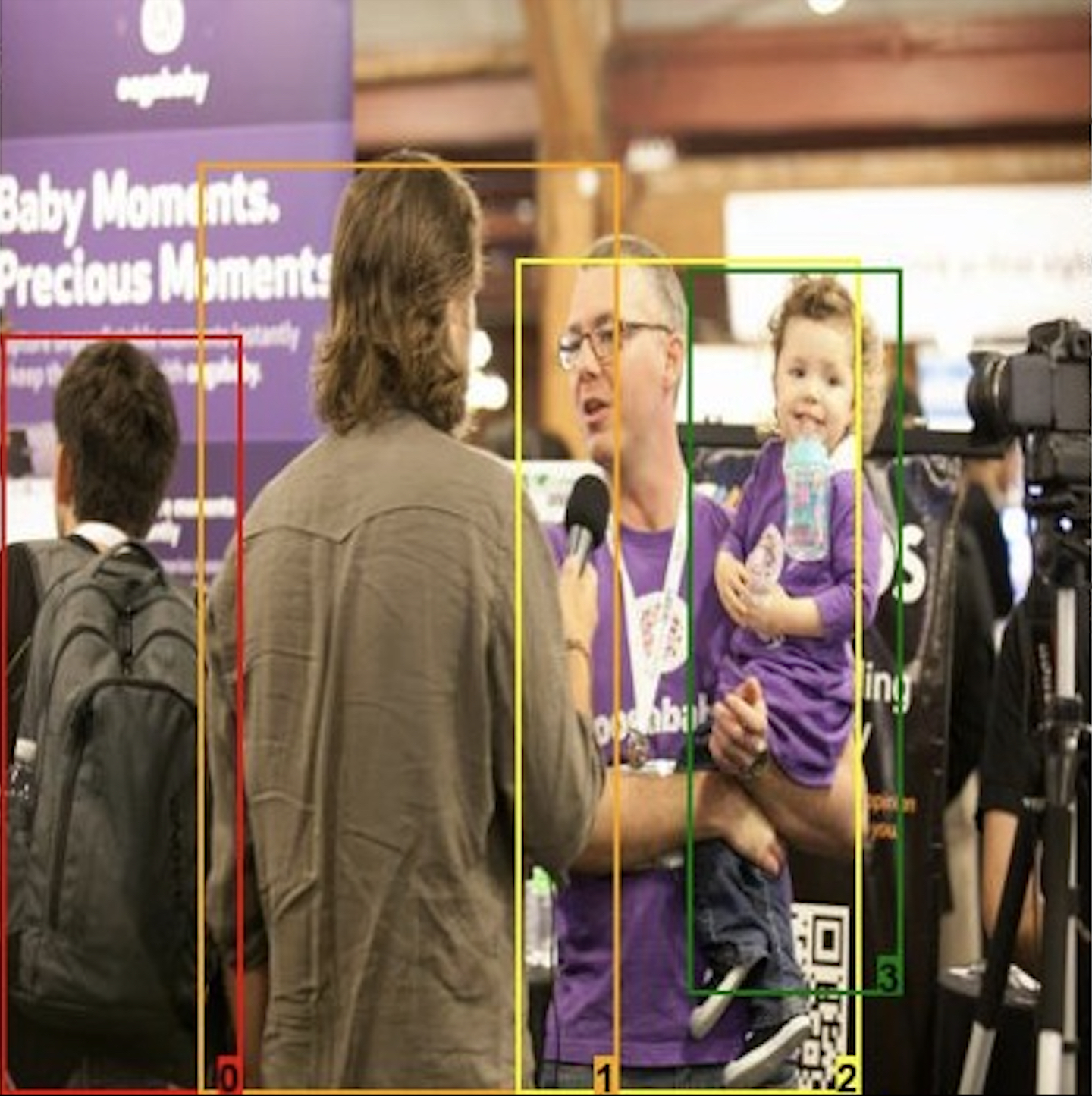} & \includegraphics[scale=0.165]{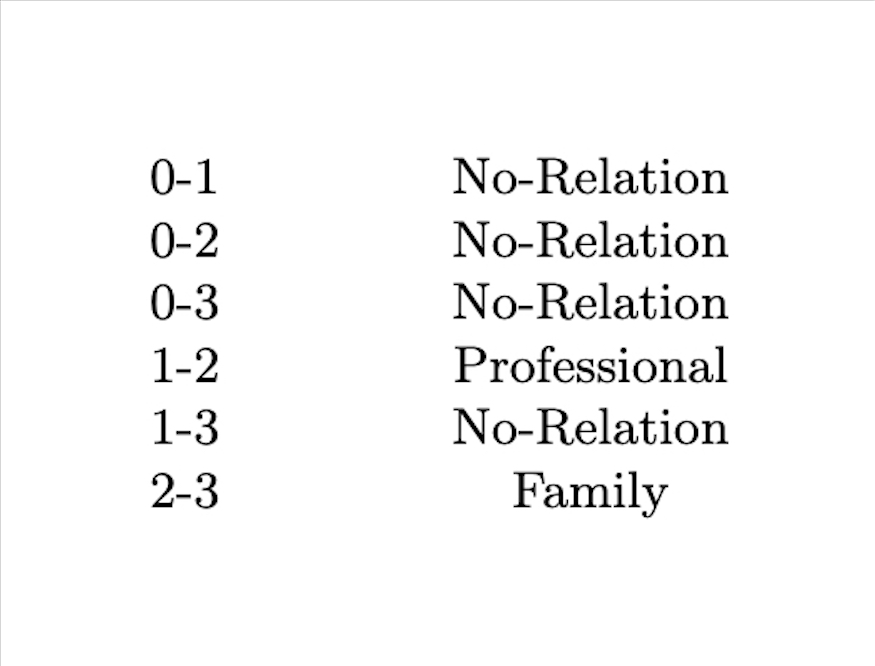} \\

\includegraphics[scale=0.0675]{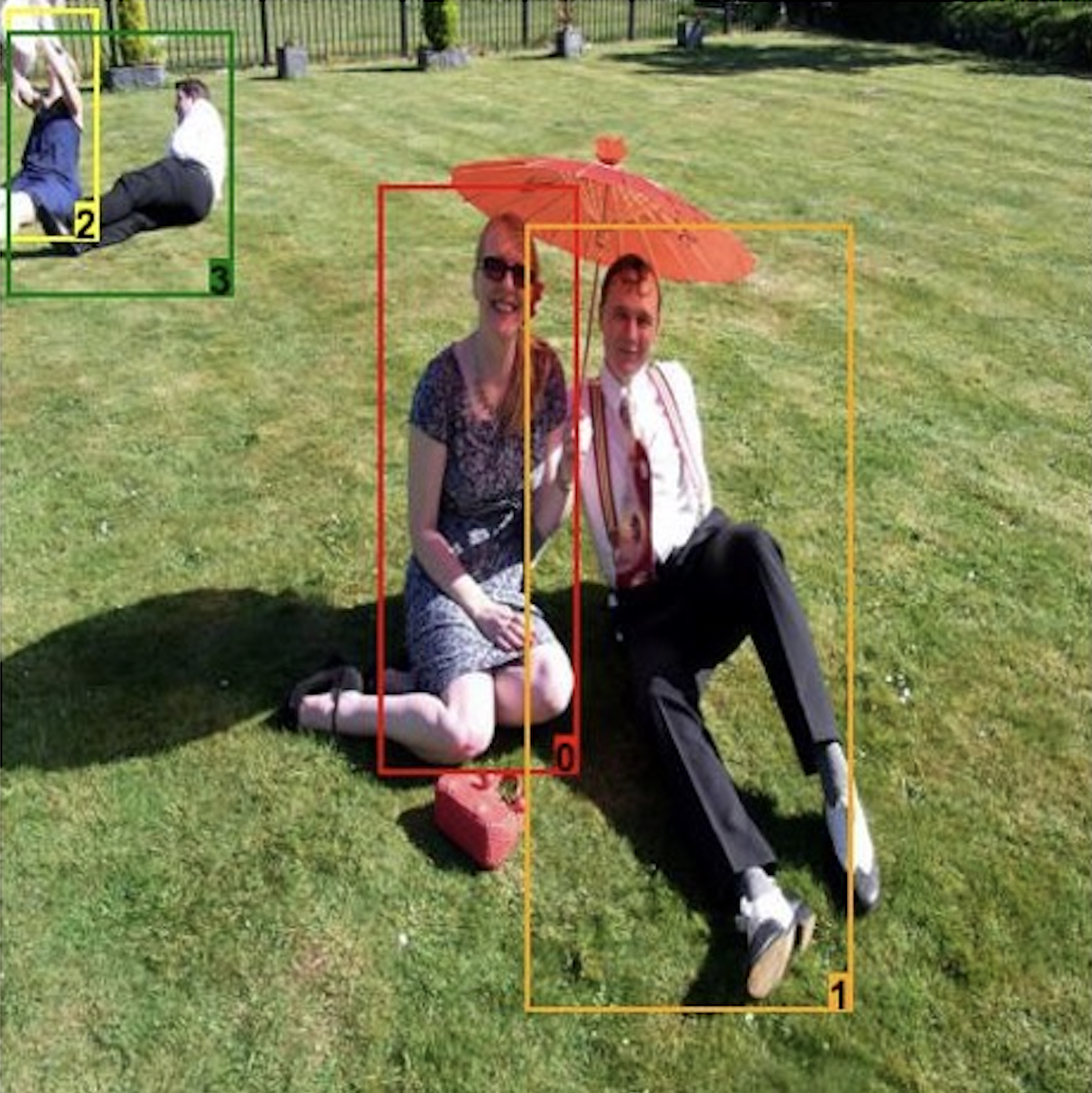} & \includegraphics[scale=0.165]{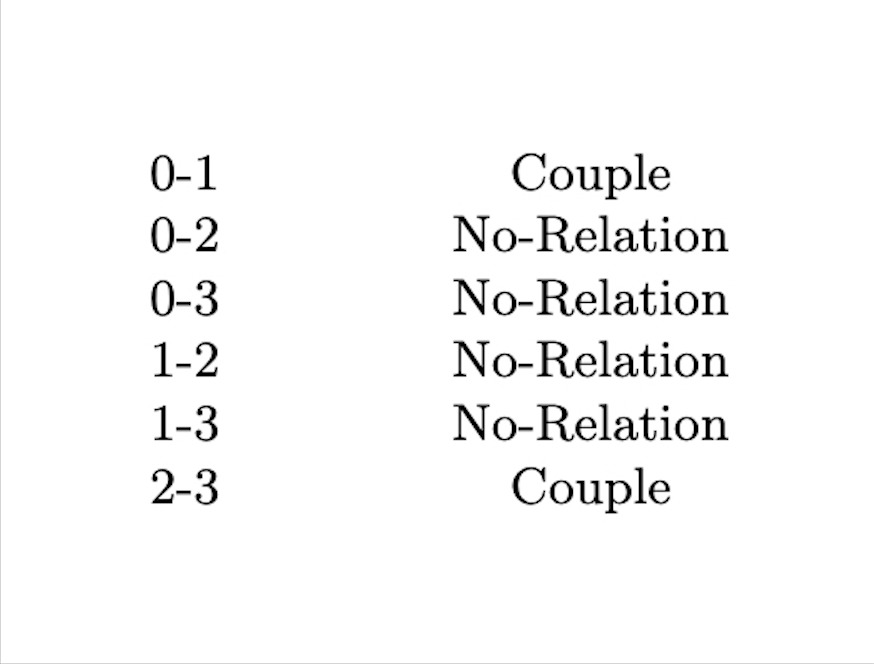} \\
\end{tabular}
\label{tab:visual_results_PISC}
\end{table}

In summary, the main contributions of this work are threefold
\begin{itemize}
  \item We analyse and improve GRIT to improve its robustness and performance. Our GRITv2 can be applied by other researchers as the base model we use is an open-source model.
  \item We also prioritize model compression, and develop GRITv2-S, which is efficiently designed and deployed on mobile using model quantization. To our best knowledge, this is the first work which also focuses on developing efficient compact models for relation recognition.
  \item Our GRITv2-L sets a new benchmark in the state-of-the-art models in interpersonal relation recognition task with PISC dataset.
\end{itemize}

\section{Related Works}
\subsection{Social Relation Recognition}
Social relation recognition in images is the task of classifying pairwise person relations using face and/or person bounding box annotations. This task is majorly benchmarked in 2 datasets from \cite{PISC_dualglance, PIPA}. The datasets are: People in Photo Albums (PIPA) \cite{PIPA} and People in Social Context (PISC) \cite{PISC_dualglance}. There have been multiple solutions\cite{PISC_dualglance, grm, mgr, srg_gn, gr2n, hf_srgr, mt_srr, isl, grit, li_et_al} to solve and improve the performance of deep learning algorithms on this task. We focus on the PISC\cite{PISC_dualglance} dataset in our work.

The pioneering work from \cite{PISC_dualglance} involved in building the PISC\cite{PISC_dualglance} dataset and providing their solution: Dual-Glance\cite{PISC_dualglance}. Dual-Glance\cite{PISC_dualglance} performs two glances on the input. The first glance generates the features of the person-pair and the union region which contains both persons. The second glance generates features of a set of region proposals in the image, which is useful in performing the relation classification. Similar to Dual-Glance\cite{PISC_dualglance}, GRM\cite{grm} also uses an object detection algorithm to generate a set of region proposals. The persons and objects are modelled as a knowledge graph, and a gated graph neural network (GGNN) \cite{ggnn} is used to predict the relations. 

In the recent years, subsequent research from \cite{cvtsrr, isl, li_et_al, trgat, grit} have proposed methods to improve the relation recognition performance to surpass state-of-the-art benchmarks. CvTSRR\cite{cvtsrr} uses the convolutional vision transformer (CvT)\cite{cvt} paired with multi-head attention mechanism\cite{transformer} to perform relation classification. TRGAT \cite{trgat} proposes a triangular reasoning based graph attention network, to improve performance of social relation reasoning. TRGAT \cite{trgat} also uses Node Contrastive Learning as a form of self-supervised learning, where an input image is augmented and contrastive learning\cite{contrastive_loss_1, contrastive_loss_2, contrastive_loss_3} is applied to make sure the features of the same person are more similar and features of different persons are more differentiated. ISL\cite{isl} uses a vision transformer \cite{vit} as an encoder to obtain person-wise features, from which relations are predicted with interpersonal similarity learning using cosine similarity between the features. ISL also introduces a new CF-loss which trains the model based on the confusion matrix obtained during the training process. \cite{li_et_al, grit} use Swin \cite{swin} as an efficient feature extraction block paired with a graph neural network (GNN) \cite{gnn} to obtain relation-level features between two people in an image. GRIT \cite{grit} also uses a transformer\cite{transformer} module to incorporate free form attention mechanism in its architecture.

\subsection{Graph Neural Networks}
Graph Neural Network (GNN) \cite{gnn} is a method applied for processing graph structured data, and Graph Convolutional networks (GCN) \cite{gcn} was first proposed to fuse convolutions with graph data, for semi-supervised learning. In the domain of Social Relation Recognition, several works \cite{gr2n, grm, mgr, trgat} adopt the GNN architecture to perform relation recognition. GRIT\cite{grit} also uses GNN to efficiently fuse global image features with localized person features. After the graph iterations, vertex features of person-pairs are concatenated to form the relation query required for relation recognition.

\begin{figure}[]
  \centering
  \includegraphics[height=6.5cm]{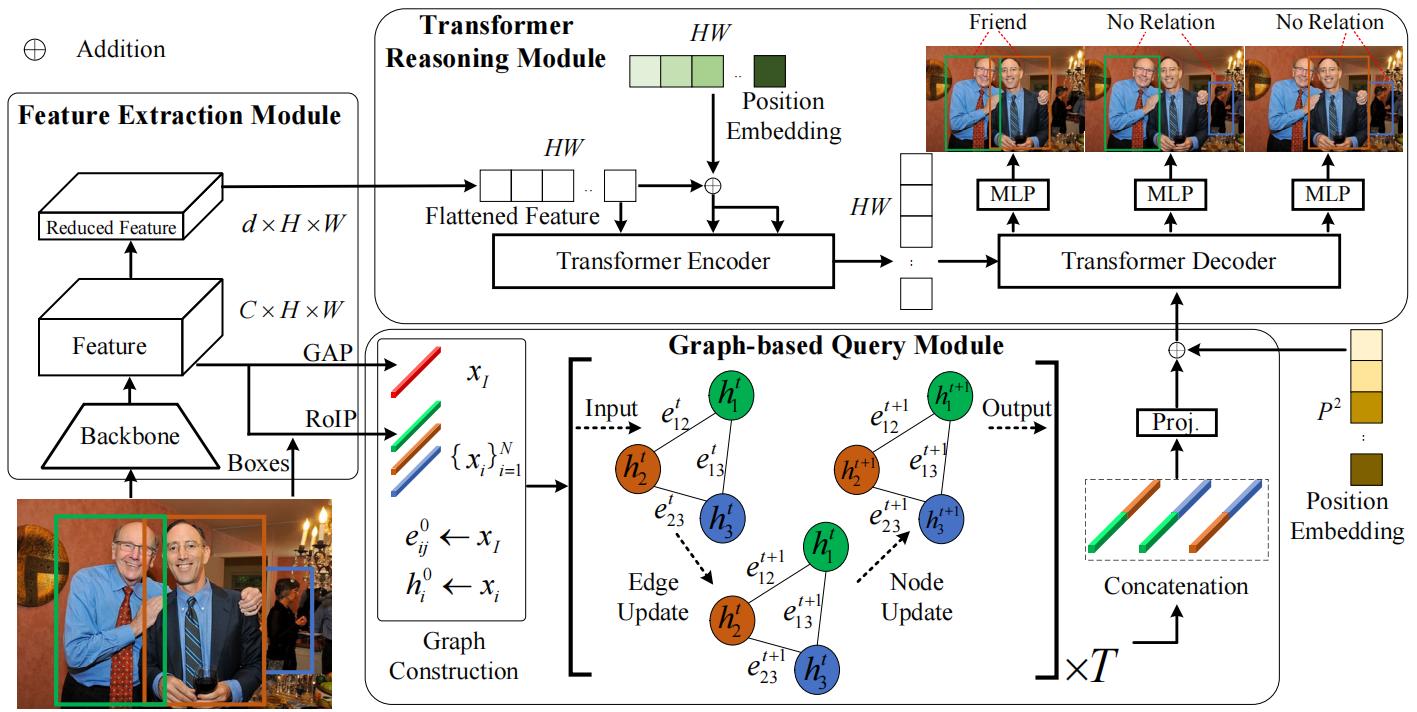}
  \caption{Architecture of GRIT\cite{grit}, consisting of 3 modules: FEM, GQM and TRM.
  }
  \label{fig:example}
\end{figure}
\begin{figure}[]
  \centering
  \includegraphics[scale=0.17]{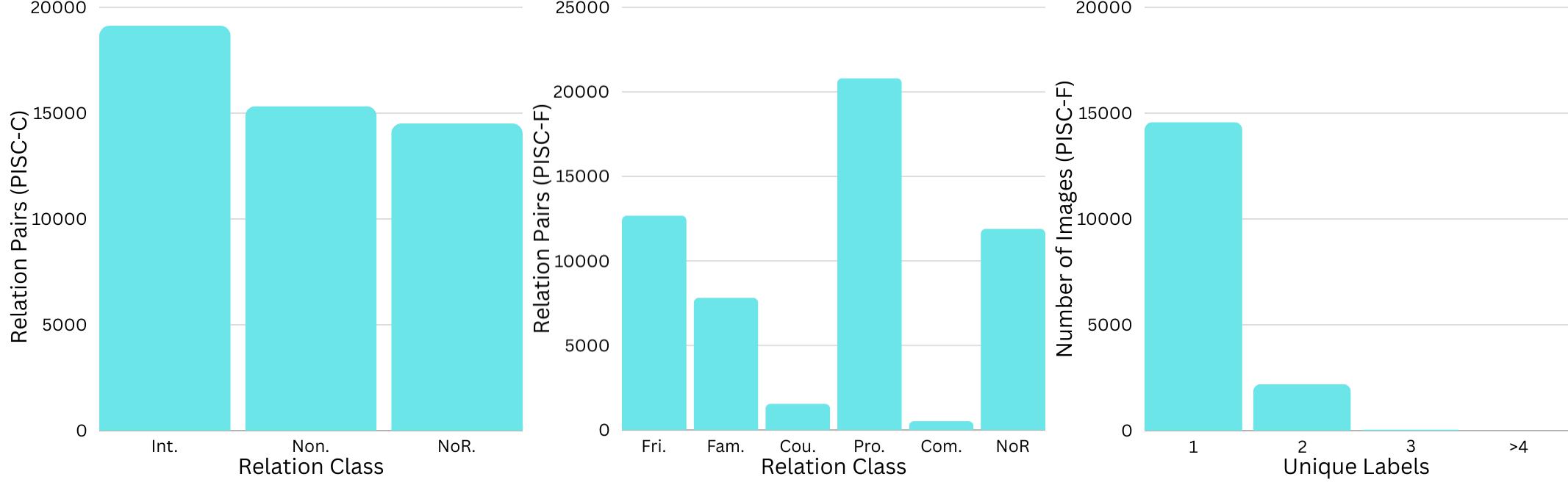}
  \caption{ PISC-C\cite{PISC_dualglance} and PISC-F\cite{PISC_dualglance} train split analysis. 
  Int: Intimate, Non: Non-Intimate, NoR: No Relation, Fri: Friend, Fam: Family, Cou: Couple, Pro: Professional, Com: Commercial.
  }
  \label{fig:pisc_train_stats}
\end{figure}
\section{GRIT}
Among the current state-of-the-art solutions, GRIT\cite{grit} stands out as the sole open-source model. The GRIT\cite{grit} model contains 3 modules, 1) Feature extraction Module (FEM) 2) Graph-based Query Module (GQM) and 3) Transformer Reasoning Module (TRM), as shown in Figure \ref{fig:example}. In this section, we provide an overview of the GRIT\cite{grit} model architecture.
\subsection{Feature Extraction Module}
The Feature Extraction Module generates global image features using powerful backbones like Swin \cite{swin}. The feature map $f_I$ obtained from the backbone is transformed into global image feature $x_I$ and local person-wise features $x_i$ using GAP(Global-Average Pooling)\cite{GAP} and  ROI(Region-of-Interest) pooling respectively (Equations \ref{equ:dt1}, \ref{equ:dt2}). The ROI Pooling is applied using the bounding boxes $b_i$ of each person detected in the image.

\begin{equation}
  x_I = GAP(f_I)
  \label{equ:dt1}
\end{equation}

\begin{equation}
  x_i = RoIP(f_I, b_i)
  \label{equ:dt2}
\end{equation}

\subsection{Graph-based Query Module}
The Graph-based Query Module has been designed to efficiently fuse local person features $x_i$ and global image features $x_I$.
The GQM is a Graph Neural Network with P nodes, where P is the maximum number of people in an image. The initial node representations are initialized to $x_i$ and all edge representations are initialized to $x_I$. The Graph Query Module performs T iterations of vertex $h_i$ updates (Equation \ref{vertex_update}) and edge $e_{ij}$ updates (Equation \ref{edge_update}). after which the relation query $q_{ij}$ between two people $i$ and $j$ is calculated by concatenating vertex features (Equation \ref{query_orig}).
\begin{equation}
  e_{ij}^{t+1} = \sigma(W^th_i^t+W^th_j^t+W_e^te_{ij}^t)
  \label{edge_update}
\end{equation}
\begin{equation}
  h_{i}^{t+1} = h_{i}^{t} + \sigma(W^th_i^t+\frac{1}{|N_i|}\sum_{j\in N_{i}}e_{ij}^{t+1} \odot W^th_j^t)
  \label{vertex_update}
\end{equation}
\begin{equation}
  q_{ij} = <h_i,h_j> 
  \label{query_orig}
\end{equation}
The variables $W^t$ and $W_{e}^t$ represent the learnable parameters of at each graph convolutional layer, with $t = 0, 1, .. T-1$ and $T$ is the number of graph iterations of node and edge updates. $\sigma$ is the RELU function, and $\odot$ is the element-wise operation. $N_i$ represents the set of neighbours of vertex i, which is the set of vertices as GRIT\cite{grit} models a complete graph in the GQM.
\subsection{Transformer Reasoning Module}
The Transformer Reasoning Module is a transformer\cite{transformer} based encoder-decoder model, implemented to visualize free-form attention in the image. The attention mechanism in the transformer allows to learn and identify important regions in an image. The global image features are encoded using the TRM encoders, and the GQM relation queries are the inputs to the TRM decoders. The number of queries to the TRM decoder are padded to $P^2$ where $P$ is the maximum number of persons per image in the dataset.


\begin{table}[tb]
  \caption{GRITv2 ablation study on PISC-F dataset\cite{PISC_dualglance} at Image resolutions 224 and 448. The column values indicate the mAP performance on the PISC-fine dataset.
  }
  \label{tab:headings-ablation-study}
  \centering
\setlength{\tabcolsep}{10pt}
  \renewcommand{\arraystretch}{1.5}
  \begin{tabular}{@{}l|llll@{}}
    \toprule

    Backbone & \multicolumn{2}{c}{TinyViT} \ & \multicolumn{2}{c}{Swin-L}\\

    \midrule
    
    Ablation Study \ & \ \ 224 \ \ \ \ \  & 448 \ \ \ & \ \ 224 \ \ \ \ \ & 448 \ \ \ \\

    \midrule
    WBCE \ & 78.85 & 81.01 & 79.96 & 83.27 \\
    + Bilateral Train,Val \ & 79.39 & 80.48 & 80.08 & 82.21 \\
    + Logit Transform \ & 79.63 & 81.16 & 80.31 & 82.46 \\
    + GQM Update \ & 80.29 & 81.64 & \textbf{80.73} & 83.30 \\
    + SE Block \ & \textbf{80.40} & \textbf{81.95} & 80.62 & \textbf{83.34}\\
  \bottomrule
  \end{tabular}
\end{table}
\section{GRITv2}

We improve GRIT\cite{grit} using GRITv2-L. We also develop GRITv2-S for mobile deployment, which is a smaller version of GRITv2. GRITv2-L uses Swin-L\cite{swin} backbone and GRITv2-S uses TinyViT-11M\cite{tiny_vit} backbone. We have performed the following improvements to GRIT\cite{grit}, as mentioned in Table \ref{tab:headings-ablation-study}.
\subsection{Weighted Binary Cross Entropy}
The relation recognition dataset contains relationship labels along with human bounding box annotations, with a relationship label corresponding to each pair of humans in each image. There exists a class imbalance in the dataset as shown in Figure \ref{fig:pisc_train_stats}. As each image consists of multiple relations, the class imbalance in the dataset cannot be directly solved by oversampling the dataset, therefore we use Weighted Binary Cross Entropy (Equations \ref{equ:wbce}, \ref{equ:wbce-weights}) using weights $w_c$ for each class.
\begin{equation}
  L = \sum_{c=1}^{C} w_cy_clogx_c
  \label{equ:wbce}
\end{equation}
\begin{equation}
  w_c = \frac{2}{n_c}\sum_{c=1}^{C}n_c
  \label{equ:wbce-weights}
\end{equation}
\subsection{Bilateral Masking}
GRIT\cite{grit} uses unilateral masking strategy, where if the relation of a person pair, A-B is labelled in the image, then the relation of the pair B-A is ignored during training and validation. This unilateral strategy is useful for training on asymmetric relations, for example: Mother-Son, Father-Son, etc. But in our experiments, we focus on the PISC\cite{PISC_dualglance} dataset which contains only symmetric labels. Therefore we use bilateral masking for better performance and robustness of GRITv2. We also evaluate the GRIT\cite{grit} model with unilateral and bilateral masking strategies in Table \ref{tab:UNILATERAL-BILATERAL}, and observe a 1-2\% performance drop in mAP, with bilateral masking. 
\begin{table}[tb]
  \caption{Performance of GRIT\cite{grit} with Bilateral masking strategy on PISC dataset\cite{PISC_dualglance}. The table consists of per-class recall for each relation and the mAP over all relations (in \%). Int: Intimate, Non: Non-Intimate, NoR: No Relation, Fri: Friend, Fam: Family,
Cou: Couple, Pro: Professional, Com: Commercial.
  }
  \label{tab:UNILATERAL-BILATERAL}
  \centering
\setlength{\tabcolsep}{3pt}
  \begin{tabular}{c|cccc|ccccccc}
    \toprule

    {} & \multicolumn{4}{c}{PISC-C} \ & \multicolumn{7}{c}{PISC-F}\\
    \midrule
    Masking Strategy & Int. & Non. & NoR & mAP & Fri. & Fam. & Cou. & Pro. & Com. & NoR & mAP \\
    \midrule
    Unilateral(default) & 84.7 & 76.8 & 70.8 & \textbf{85.6} & 70.6  &  71.9  &  55.5  &  87.7  &  25.7  &   78.3  & \textbf{78.3}\\
    Bilateral & 85.1 & 77.2 & 65.7 & \textbf{84.3} & 71.3  &  73.0  &  45.7  &  88.4  &  24.3  &  74.0  & \textbf{76.2}\\

  \bottomrule
  \end{tabular}
\end{table}

\subsection{Logit Transformation}
From section 4.2, we have discussed and introduced bilateral masking in the dataset for improving the performance and robustness of GRITv2. We also make a change in the GRIT\cite{grit} model, by making a logit-level transformation which ensures that symmetric nature in relation-prediction is followed by the model. Assuming there are P people in an image, the final layer performs classification on a logit matrix $\textit{m} \in \mathbb{R}^{PXP}$. We apply a transformation on the logit matrix, by adding it with its transpose, to make it symmetric (Equation \ref{equ:dt3}). This approach ensures the model's predictions to be symmetric.
\begin{equation}
  m = m + m^T
  \label{equ:dt3}
\end{equation}
\subsection{GQM Update}
 GRIT\cite{grit} uses a Graph Convolutional Neural Network to get the pairwise relation representation in an image. The vertices represent each person in the image, and the edges are represent the message-passing between the vertices. After the graph iterations, the vertex features of each of the 2 people in a relation are concatenated and used for Relation classification. 

We use the edge features $e_{ij}$ to represent the relationship features between two nodes, this approach also reduces the dimension of the resulting feature embeddings as feature concatenation is not required. This approach consumes lesser memory and also more efficient in improving the performance of the model. The relation query $q_{ij}$ in GRIT\cite{grit} is initialized with the concatenation of vertex features $h_i$ and $h_j$, but GRITv2 uses the graph edge $e_{ij}$ as the relation query.(Equation \ref{equ:GQM})

\begin{equation}
  q_{ij} = e_{ij}
  \label{equ:GQM}
\end{equation}

\subsection{Squeeze and Excitation Block}
After including the previous changes, there still exists a performance gap between GRITv2-L and GRITv2-S (Table \ref{tab:headings-ablation-study}). We introduce a Squeeze and Excitation \cite{se_block} block after the Feature Extraction Module, for the model to improve and recalibrate the feature representations $f_I$ by increasing importance to informative features/regions and decreasing the focus on less useful ones (Equation \ref{equ:SE}). 

\begin{equation}
  f_I = SE(f_I)
  \label{equ:SE}
\end{equation}

\subsection{Comparision with SOTA}
In this section we compare GRITv2 with existing state-of-the-art models, in Table \ref{tab:SOTA}. From the result comparison in Table \ref{tab:SOTA}, we can see that GRITv2-L outperforms the existing benchmarks in both PISC-F and PISC-C relation recognition datasets. GRITv2-S is within $2\%$ performance gap of GRITv2-L with just 0.0625x the model size and parameters.

We also compare GRITv2 with TRGAT-NCL\cite{trgat} and CvTSRR\cite{cvtsrr} in Table \ref{tab:2023-SOTA}. We have compiled this comparison table separately as the training and evaluation conditions of these models are different compared to remaining methods.

\subsubsection{TRGAT-NCL:}
TRGAT-NCL\cite{trgat} uses a randomly sampled training resolution between  600, 720 and 960. The inference/testing resolution is set to 720, therefore we upscale GRITv2 to the resolution of 448 and compare for fair comparison. 

\subsubsection{CvTSRR:}
CvTSRR\cite{cvtsrr} uses a custom Train-Val-Test split of 80-10-10, which differs from the conventional PISC relation splits for the coarse and fine tasks. The conventional Train-Val-Test splits for the PISC-C task is 62-18-20 and the Train-Val-Test split for the PISC-F task is 91-3-6. This explains the huge improvement made by CvTSRR in the PISC-C relation task compared to the PISC-F relation task.

\begin{table}[tb]
  \caption{Comparison of GRITv2 with state-of-the-art methods evaluated on the PISC dataset\cite{PISC_dualglance}. The top-2 best performing methods based on mAP are highlighted. The table consists of per-class recall for each relation and the mAP over all relations (in \%). Int: Intimate, Non: Non-Intimate, NoR: No Relation, Fri: Friend, Fam: Family,
Cou: Couple, Pro: Professional, Com: Commercial.
  }
  \label{tab:SOTA}
  \centering
\setlength{\tabcolsep}{2.25pt}
  \begin{tabular}{c|cccc|ccccccc}
    \toprule

    {} & \multicolumn{4}{c}{PISC-C} \ & \multicolumn{7}{c}{PISC-F}\\
    \midrule
    Model & Int. & Non. & NoR & mAP & Fri. & Fam. & Cou. & Pro. & Com. & NoR & mAP \\
    \midrule
    Dual-Glance\cite{PISC_dualglance}& 73.1 & 84.2 & 59.6 & {79.7} & 35.4  &  68.1  &  76.3  &  70.3  &  57.6  &  60.9  & {63.2}\\
    GRM \cite{grm} & 81.7 & 73.4 & 65.5 & {82.8} & 59.6  &  64.4  &  58.6  &  76.6  &  39.5  &   67.7  & {68.7}\\
    MGR \cite{mgr} & - & - & - & - & 64.6  &  67.8  &  60.5  &  76.8  &  34.7  &  70.4  & {70.0}\\
    SRG-GN \cite{srg_gn}& - & - & - & - & 25.2  &  80.0  &  100.0  &  78.4  &  83.3  &   62.5  & {71.6}\\
    GR2N \cite{gr2n}& 81.6 & 74.3 & 70.8 & {83.1} & 60.8  &  65.9 & 84.8  &  73.9  &  51.7  &  70.4  & {72.7}\\
    HF-SRGR\cite{hf_srgr} & 89.1 & 87.0 & 75.5 & {84.6} & 82.2  &  39.4  &  33.2  &  60.0  &  47.7  &   71.8  & {73.0}\\
    MT-SRR\cite{mt_srr} & 91.8 & 91.8 & 75.2 & {86.8} & 71.6  &  69.7  &  62.5  &  88.0  &  34.2  &  72.7  & {74.6}\\
    ISL\cite{isl} & 92.8 & 91.6 & 75.7 & \textbf{87.0} & 78.2  &  72.6  &  70.0  &  88.6  &  42.9  &   81.1  & {75.6}\\
    GRIT (Uni)\cite{grit} & 84.7 & 76.8 & 70.8 & {85.6} & 70.6  &  71.9  &  55.5  &  87.7  &  25.7  &   78.3  & {78.3}\\
    GRIT (Bi)\cite{grit} & 85.1 & 77.2 & 65.7 & {84.3} & 71.3  &  73.0  &  45.7  &  88.4  &  24.3  &  74.0  & {76.2} \\ 
    (Li et al. 2022)\cite{li_et_al} & 86.3 & 78.5 & 71.4 & {86.8} & 72.2  &  74.0  &  81.2  &  82.5  &  60.4  &  70.2  & {79.6} \\ 
    \midrule
    GRITv2-L (224) & 86.59 & 75.38 & 76.43 & {\textbf{87.83}} & 71.01 & 79.47 & 78.52 & 86.13 & 61.58 & 73.75 & \textbf{80.73} \\
    GRITv2-S (224) & 78.91 & 79.42 & 75.94 & {85.89} & 75.82  &  72.38  &  60.16  &  83.43  &  40.40  &  78.87  & \textbf{80.40}\\
    \midrule

  \bottomrule
  \end{tabular}
\end{table}
\begin{table}[tb]
  \caption{Comparison of GRITv2 with TRGAT-NCL and CvTSRR on PISC dataset\cite{PISC_dualglance}. The top-2 best performing methods based on mAP are highlighted. The table consists of per-class recall for each relation and the mAP
over all relations (in \%). Int: Intimate, Non: Non-Intimate, NoR: No Relation, Fri:
Friend, Fam: Family, Cou: Couple, Pro: Professional, Com: Commercial.
  }
  \label{tab:2023-SOTA}
  \centering
\setlength{\tabcolsep}{2.25pt}
  \begin{tabular}{c|cccc|ccccccc}
    \toprule

    {} & \multicolumn{4}{c}{PISC-C} \ & \multicolumn{7}{c}{PISC-F}\\
    \midrule
    Model & Int. & Non. & NoR & mAP & Fri. & Fam. & Cou. & Pro. & Com. & NoR & mAP \\
    \midrule
    TRGAT-NCL & 81.4 & 76.6 & 75.3 & {87.6} & 58.2  &  73.1  &  78.9  &  76.7  &  70.6  &   73.6  & {78.2}\\
    CvTSRR & 89.68 & 70.79 & 83.00 & \textbf{90.25} & 76.90  &  75.00  &  32.10  &  78.93  &  44.23  &  84.01  & {75.32} \\ 
    \midrule
    GRITv2-L (448) & 87.12 & 79.75 & 77.75 & \textbf{89.81} & 80.38 & 74.00 & 77.73 & 84.37 & 58.47 & 77.67 & \textbf{83.34} \\
    GRITv2-L (224) & 86.59 & 75.38 & 76.43 & {{87.83}} & 71.01 & 79.47 & 78.52 & 86.13 & 61.58 & 73.75 & {80.73} \\
    GRITv2-S (448) & 83.26 & 81.05 & 70.40 & {88.03} & 70.38 & 75.78 & 83.59 & 83.67 & 62.43 & 67.43 & \textbf{81.95} \\
    GRITv2-S (224) & 78.91 & 79.42 & 75.94 & {85.89} & 75.82  &  72.38  &  60.16  &  83.43  &  40.40  &  78.87  & {80.40}\\
  \bottomrule
  \end{tabular}
\end{table}

\section{Experiments}
\subsection{Datasets}
There are two widely used social relation recognition datasets, PIPA\cite{PIPA} and PISC\cite{PISC_dualglance}. We have performed all of our experiments on the PISC dataset as the PIPA dataset is only partially available using the Flickr API\cite{flickr}. We have tried recovering the dataset, but we have only been able to recover upto 80\% of the PIPA dataset.

The PISC dataset containes 22670 images, with 2 relation recognition splits, PISC-Coarse and PISC-Fine. The coarse-level relations consists of 3 relations (intimate, non-intimate, no relation) and the fine-level consists of 6 relations (friend, family, couple, professional, commercial, and no relation). Following the previous methods on PISC dataset, we use per-class recalls and mean Average Precision (mAP) to evaluate model performance. 

We also visualize number of unique relations per image in the PISC-F train split(Figure \ref{fig:pisc_train_stats}) and identify the need to improve relation recognition datasets by collecting images with diverse relations between people.
\subsection{Implementation Details}
Similar to GRIT\cite{grit}, we set the learning rate of the backbone and the rest of the network to be $10^{-5}$ and $10^{-4}$ respectively. We use a batch size of 12 for GRITv2-S and a batch size of 4 for GRITv2-L respectively. The number of layers in GQM is 2. The TRM module is trained with a dropout of 0.2 with 1 encoder layer and 1 decoder layer with 8 heads each. We conduct all experiments on a single NVIDIA V100 GPU.

\subsection{Model Compression}
We observe that the backbone of the model contains the most parameters compared to other modules in GRITv2 (Table \ref{tab:Params}). We follow 2 approaches to compress the GRITv2 model, 1) MiniViT\cite{minivit} distillation, 2) Backbone Selection.
\begin{table}[tb]
  \caption{Parameter comparison between GRITv2-L and GRITv2-S. The backbone (FEM module) consists of the majority of parameters in GRITv2.
  }
  \label{tab:Params}
  \centering
\setlength{\tabcolsep}{3pt}
  \begin{tabular}{c|cc}
    \toprule

    {} & \multicolumn{2}{c}{Parameters} \\
    \midrule
    Model & Backbone & Total \\
    \midrule
    GRITv2-L & 197M & 242M\\
    GRITv2-S & 11M & 15.1M\\
    \midrule

  \bottomrule
  \end{tabular}
\end{table}
\begin{table}[tb]
  \caption{Mini-Swin knowledge distillation experiment on PISC-F dataset\cite{PISC_dualglance}.
  }
  \label{Minivit_exp}
  \centering
\setlength{\tabcolsep}{3pt}
  \begin{tabular}{c|cc}
    \toprule
    Backbone & Parameters & mAP(PISC-F) \\
    \midrule
    Swin-T & 28M & 77.31 \\
    Mini-Swin-T & 12M & 70.12 \\
    Mini-Swin-T(distill) & 12M & 74.06 \\
    \midrule

  \bottomrule
  \end{tabular}
\end{table}
\subsubsection{MiniViT Distillation:} 
MiniViT\cite{minivit} is a weight-sharing and weight-multiplexing based relational knowledge distillation strategy. The MiniViT\cite{minivit} strategy provided Mini-Swin models which outperform the teacher model in the ImageNet \cite{imagenet} classification task, while having around 50\% of the teacher's parameter count. We consider Mini-Swin-T, and perform MiniViT\cite{minivit} distillation experiment (Table \ref{Minivit_exp}) to observe improvement over normal training experiments on the PISC-F dataset.
\subsubsection{Backbone selection:} 
We select TinyViT\cite{tiny_vit} backbone, as it adopts a heirarchical vision transformer architecture similar to Swin\cite{swin}, and generates multi-scale image features. We also give preference to TinyViT\cite{tiny_vit} due to the vast pretraining distillation performed on the TinyViT\cite{tiny_vit} student models on the ImageNet-21k\cite{imagenet} dataset. 
\begin{table}[tb]
  \caption{Performance of GRITv2-S before and after Quantization Aware Training PISC dataset\cite{PISC_dualglance}. The table consists of per-class recall for each relation and the mAP over all relations (in \%). Int: Intimate, Non: Non-Intimate, NoR: No Relation, Fri: Friend, Fam: Family,
Cou: Couple, Pro: Professional, Com: Commercial.
  }
  \label{tab:QAT}
  \centering
\setlength{\tabcolsep}{1.75pt}
  \begin{tabular}{c|cccc|ccccccc}
    \toprule

    {} & \multicolumn{4}{c}{PISC-C} \ & \multicolumn{7}{c}{PISC-F}\\
    \midrule
    Masking Strategy & Int. & Non. & NoR & mAP & Fri. & Fam. & Cou. & Pro. & Com. & NoR & mAP \\
    \midrule
    GRITv2-S (FP32) & 78.91 & 79.42 & 75.94 & \textbf{85.89} & 75.82  &  72.38  &  60.16  &  83.43  &  40.40  &  78.87  & \textbf{80.40} \\
    GRITv2-S (INT8) & 81.13 & 76.97 & 74.46 & \textbf{85.37} & 73.58  &  74.23  &  65.81  &  81.93  &  38.85  &  79.42  & \textbf{80.14} \\
    \midrule

  \bottomrule
  \end{tabular}
\end{table}
\subsection{Quantization}
  GRITv2-S, initially sized at 64 MB, underwent Quantization Aware Training (QAT) \cite{QAT} to efficiently reduce its size to 22 MB with minimal performance loss (Table \ref{tab:QAT}). By converting the model from floating-point (FP32) precision to 8-bit integer (Int8) precision through QAT, we successfully optimized it for deployment on the flagship OnePlus 12 device (Adreno 750 GPU). Upon deployment, we measured a latency of 683 milliseconds. This model also outperforms previous state-of-the-art benchmarks on the PISC-F dataset. Our approach demonstrates the effectiveness of quantization techniques in reducing model size, enabling integration of deep learning models into resource-constrained devices.
\section{Conclusion}
We propose GRITv2, a new state-of-the-art model which surpasses existing state-of-the-art models in PISC\cite{PISC_dualglance} relation recognition task. The proposed GRITv2-S is an efficient light-weight model, with a model size of 22MB during model deployment and still superior to existing PISC-F benchmarks. Our research work has better applicabilty as it can be replicated and applied using GRIT\cite{grit}. We also provide our approach in performance enhancement and model compression to deploy efficient models on mobile devices. As we establish a new state-of-the-art benchmark in the PISC relation dataset, we plan to analyse and improve other relation recognition tasks as a future research direction.


%
%

\bibliographystyle{splncs04}
\bibliography{main}

\begin{thebibliography}{10}
\providecommand{\url}[1]{\texttt{#1}}
\providecommand{\urlprefix}{URL }
\providecommand{\doi}[1]{https://doi.org/#1}

\bibitem{flickr}
flickr: \url{https://www.flickr.com/}

\bibitem{grit}
grit: \url{https://github.com/IFBigData/GRIT/}

\bibitem{QAT}
Bondarenko, Y., Nagel, M., Blankevoort, T.: Understanding and overcoming the challenges of efficient transformer quantization. ArXiv  \textbf{abs/2109.12948} (2021), \url{https://api.semanticscholar.org/CorpusID:237940329}

\bibitem{contrastive_loss_2}
Dai, B., Lin, D.: Contrastive learning for image captioning. In: Neural Information Processing Systems (2017), \url{https://api.semanticscholar.org/CorpusID:7900381}

\bibitem{imagenet}
Deng, J., Dong, W., Socher, R., Li, L.J., Li, K., Fei-Fei, L.: Imagenet: A large-scale hierarchical image database. In: 2009 IEEE Conference on Computer Vision and Pattern Recognition. pp. 248--255 (2009). \doi{10.1109/CVPR.2009.5206848}

\bibitem{vit}
Dosovitskiy, A., Beyer, L., Kolesnikov, A., Weissenborn, D., Zhai, X., Unterthiner, T., Dehghani, M., Minderer, M., Heigold, G., Gelly, S., Uszkoreit, J., Houlsby, N.: An image is worth 16x16 words: Transformers for image recognition at scale. ICLR  (2021)

\bibitem{srg_gn}
Goel, A., Ma, K.T., Tan, C.: An end-to-end network for generating social relationship graphs. 2019 IEEE/CVF Conference on Computer Vision and Pattern Recognition (CVPR) pp. 11178--11187 (2019), \url{https://api.semanticscholar.org/CorpusID:85498036}

\bibitem{trgat}
Guo, Y., Yin, F., Feng, W., Yan, X., Xue, T., Mei, S., Liu, C.L.: Social relation reasoning based on triangular constraints. Proceedings of the AAAI Conference on Artificial Intelligence  \textbf{37}(1),  737--745 (Jun 2023). \doi{10.1609/aaai.v37i1.25151}, \url{https://ojs.aaai.org/index.php/AAAI/article/view/25151}

\bibitem{contrastive_loss_1}
Hu, H., Cui, J., Wang, L.: Region-aware contrastive learning for semantic segmentation. In: 2021 IEEE/CVF International Conference on Computer Vision (ICCV). pp. 16271--16281 (2021). \doi{10.1109/ICCV48922.2021.01598}

\bibitem{se_block}
Hu, J., Shen, L., Sun, G.: Squeeze-and-excitation networks. In: 2018 IEEE/CVF Conference on Computer Vision and Pattern Recognition. pp. 7132--7141 (2018). \doi{10.1109/CVPR.2018.00745}

\bibitem{gcn}
Kipf, T., Welling, M.: Semi-supervised classification with graph convolutional networks  (09 2016)

\bibitem{li_et_al}
Li, H., Chen, N., Jiang, Y.: Social relationship recognition based on shifted windows transformer and dynamic attention graph neural network. pp. 300--305 (08 2022). \doi{10.1109/MLISE57402.2022.00066}

\bibitem{PISC_dualglance}
Li, J., Wong, Y., Zhao, Q., Kankanhalli, M.: Dual-glance model for deciphering social relationships. 2017 IEEE International Conference on Computer Vision (ICCV) pp. 2669--2678 (2017), \url{https://api.semanticscholar.org/CorpusID:10027965}

\bibitem{hf_srgr}
Li, L., Qing, L., Wang, Y., Su, J., Cheng, Y., Peng, Y.: Hf-srgr: a new hybrid feature-driven social relation graph reasoning model. The Visual Computer  \textbf{38} (07 2021). \doi{10.1007/s00371-021-02244-w}

\bibitem{gr2n}
Li, W., Duan, Y., Lu, J., Feng, J., Zhou, J.: Graph-based social relation reasoning. In: ECCV (2020)

\bibitem{ggnn}
Li, Y., Tarlow, D., Brockschmidt, M., Zemel, R.S.: Gated graph sequence neural networks. In: Bengio, Y., LeCun, Y. (eds.) 4th International Conference on Learning Representations, {ICLR} 2016, San Juan, Puerto Rico, May 2-4, 2016, Conference Track Proceedings (2016), \url{http://arxiv.org/abs/1511.05493}

\bibitem{GAP}
Lin, M., Chen, Q., Yan, S.: Network in network. CoRR  \textbf{abs/1312.4400} (2013), \url{https://api.semanticscholar.org/CorpusID:16636683}

\bibitem{swin}
Liu, Z., Lin, Y., Cao, Y., Hu, H., Wei, Y., Zhang, Z., Lin, S., Guo, B.: Swin transformer: Hierarchical vision transformer using shifted windows. In: Proceedings of the IEEE/CVF International Conference on Computer Vision (ICCV) (2021)

\bibitem{gnn}
Scarselli, F., Gori, M., Tsoi, A.C., Hagenbuchner, M., Monfardini, G.: The graph neural network model. IEEE Transactions on Neural Networks  \textbf{20}(1),  61--80 (2009). \doi{10.1109/TNN.2008.2005605}

\bibitem{cvtsrr}
Shultana, S., Li, L., Li, N.: Cvtsrr: A convolutional vision transformer based method for social relation recognition. In: 2023 International Conference on Communications, Computing and Artificial Intelligence (CCCAI). pp. 64--68. IEEE Computer Society, Los Alamitos, CA, USA (jun 2023). \doi{10.1109/CCCAI59026.2023.00020}, \url{https://doi.ieeecomputersociety.org/10.1109/CCCAI59026.2023.00020}

\bibitem{isl}
Tang, W., Qing, L., Gou, H., Guo, L., Peng, Y.: Unveiling social relations: Leveraging interpersonal similarity learning for social relation recognition. IEEE Signal Processing Letters  \textbf{PP}, ~1--5 (01 2023). \doi{10.1109/LSP.2023.3306152}

\bibitem{transformer}
Vaswani, A., Shazeer, N., Parmar, N., Uszkoreit, J., Jones, L., Gomez, A.N., Kaiser, L.u., Polosukhin, I.: Attention is all you need. In: Guyon, I., Luxburg, U.V., Bengio, S., Wallach, H., Fergus, R., Vishwanathan, S., Garnett, R. (eds.) Advances in Neural Information Processing Systems. vol.~30. Curran Associates, Inc. (2017), \url{https://proceedings.neurips.cc/paper_files/paper/2017/file/3f5ee243547dee91fbd053c1c4a845aa-Paper.pdf}

\bibitem{mt_srr}
Wang, Y., Qing, L., Wang, Z., Cheng, Y., Peng, Y.: Multi-level transformer-based social relation recognition. Sensors  \textbf{22}(15) (2022). \doi{10.3390/s22155749}, \url{https://www.mdpi.com/1424-8220/22/15/5749}

\bibitem{grm}
Wang, Z., Chen, T., Ren, J.S.J., Yu, W., Cheng, H., Lin, L.: Deep reasoning with knowledge graph for social relationship understanding. In: International Joint Conference on Artificial Intelligence (2018), \url{https://api.semanticscholar.org/CorpusID:49558620}

\bibitem{cvt}
Wu, H., Xiao, B., Codella, N., Liu, M., Dai, X., Yuan, L., Zhang, L.: Cvt: Introducing convolutions to vision transformers. In: 2021 IEEE/CVF International Conference on Computer Vision (ICCV). pp. 22--31 (2021). \doi{10.1109/ICCV48922.2021.00009}

\bibitem{tiny_vit}
Wu, K., Zhang, J., Peng, H., Liu, M., Xiao, B., Fu, J., Yuan, L.: Tinyvit: Fast pretraining distillation for small vision transformers. In: European conference on computer vision (ECCV) (2022)

\bibitem{contrastive_loss_3}
Wu, Z., Xiong, Y., Yu, S.X., Lin, D.: Unsupervised feature learning via non-parametric instance discrimination. In: 2018 IEEE/CVF Conference on Computer Vision and Pattern Recognition. pp. 3733--3742 (2018). \doi{10.1109/CVPR.2018.00393}

\bibitem{minivit}
Zhang, J., Peng, H., Wu, K., Liu, M., Xiao, B., Fu, J., Yuan, L.: Minivit: Compressing vision transformers with weight multiplexing. In: Proceedings of the IEEE/CVF Conference on Computer Vision and Pattern Recognition (CVPR). pp. 12145--12154 (June 2022)

\bibitem{mgr}
Zhang, M., Liu, X., Liu, W., Zhou, A., Ma, H., Mei, T.: Multi-granularity reasoning for social relation recognition from images. 2019 IEEE International Conference on Multimedia and Expo (ICME) pp. 1618--1623 (2019), \url{https://api.semanticscholar.org/CorpusID:57761184}

\bibitem{PIPA}
Zhang, N., Paluri, M., Taigman, Y., Fergus, R., Bourdev, L.D.: Beyond frontal faces: Improving person recognition using multiple cues. 2015 IEEE Conference on Computer Vision and Pattern Recognition (CVPR) pp. 4804--4813 (2015), \url{https://api.semanticscholar.org/CorpusID:13313217}

\end{thebibliography}
\end{document}